\pdfoutput=1

\documentclass[11pt]{article}

\usepackage[preprint]{acl}

\usepackage{times}
\usepackage{latexsym}

\usepackage[T1]{fontenc}

\usepackage[utf8]{inputenc}

\usepackage{microtype}

\usepackage{inconsolata}

\usepackage{graphicx}

\usepackage{booktabs}

\usepackage {lscape}
\usepackage{pdflscape}
\usepackage{rotating}
\usepackage{multirow}

\title{Do LLMs Implicitly Determine the Suitable Text Difficulty  for Users?}

\author{Author 1 \and ... \and Author n \\
         Address line \\ ... \\ Address line}

\author{Seiji Gobara, Hidetaka Kamigaito, Taro Watanabe \\
        Nara Institute of Science and Technology  \\
        \texttt{\{seiji.gobara.gt6, kamigaito.h, taro\}@is.naist.jp}}

\begin{document}
\maketitle
\begin{abstract}
Education that suits the individual learning level is necessary to improve students' understanding. The first step in achieving this purpose by using large language models (LLMs) is to adjust the textual difficulty of the response to students.
This work analyzes how LLMs can implicitly adjust text difficulty between user input and its generated text. To conduct the experiments, we created a new dataset from Stack-Overflow to explore the performance of question-answering-based conversation.
Experimental results on the Stack-Overflow dataset and the TSCC dataset, including multi-turn conversation show that LLMs can implicitly handle text difficulty between user input and its generated response. We also observed that some LLMs can surpass humans in handling text difficulty and the importance of instruction-tuning.
\end{abstract}

\section{Introduction}
Following the advance of Large Language Models (LLMs), educational applications start to leverage LLMs. 
\newcite{dijkstra2022reading} use LLMs to spark curiosity for boosting children's motivation to learn. \newcite{gabajiwala2022quiz}  incorporate LLMs into interactive elements such as quizzes and flashcards to enhance engagement and learning of users. 

Besides, LLMs play a crucial role in text simplification, a task to transform complex text into simpler one with keeping original meaning. Due to the characteristic, text simplification can make educational content more accessible \cite{al2021automated}. \newcite{feng2023sentence} utilize LLMs such as GPT-3.5 \cite{ouyang2022training} for both zero-shot and few-shot text simplification and \newcite{rooein2023know} adjust text difficulty by LLMs.

\begin{figure}[t]
  \centering
\includegraphics[width=0.95\columnwidth]{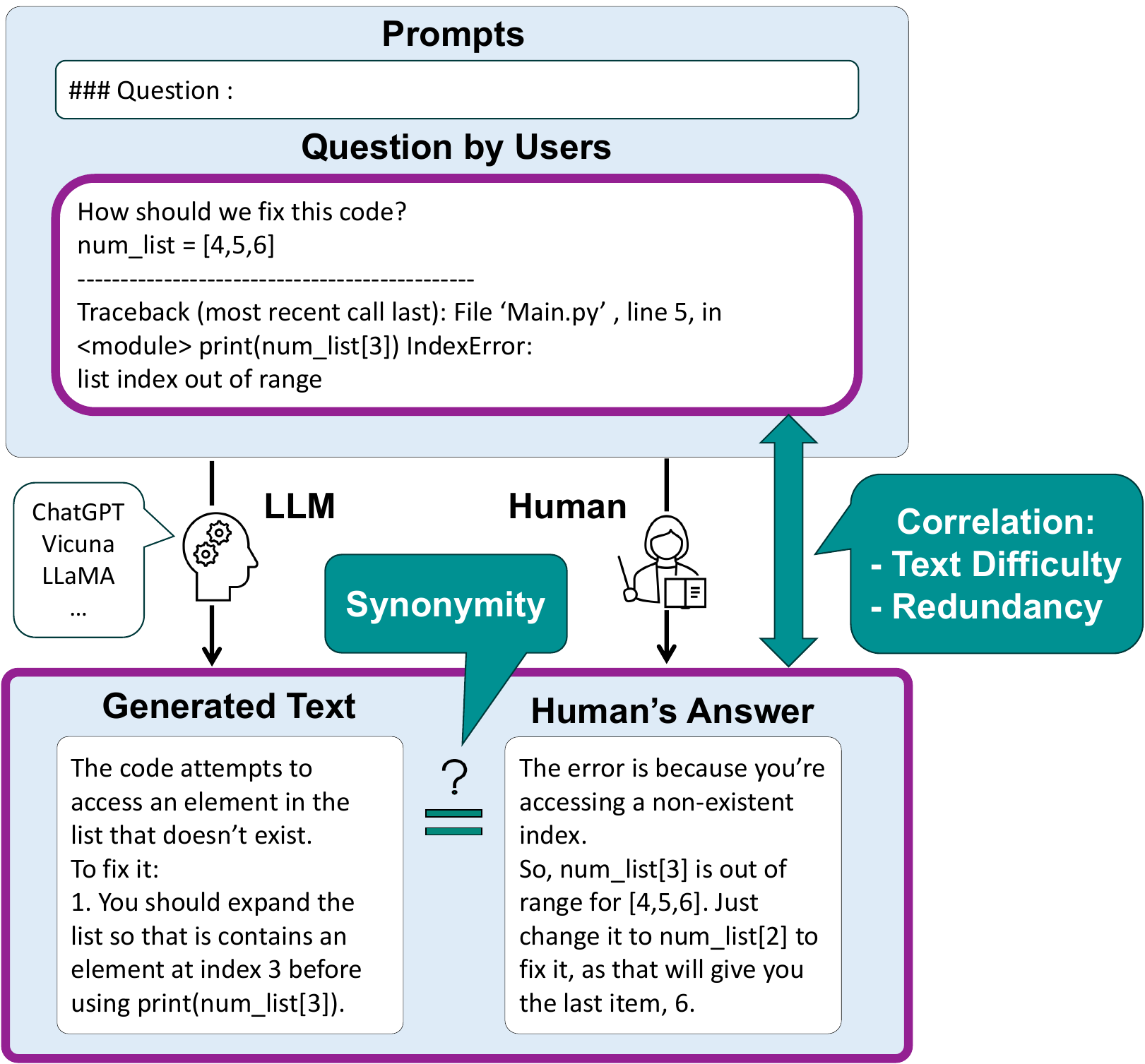}
  \caption{Overview of our evaluation procedure. We evaluate generated texts from LLMs for user questions by comparing the correlation of text difficulty and redundancy. We also evaluate the synonymity between generated texts by LLMs and human answers.}
  \label{fig:overview}
\end{figure}
Even if text simplification can support education, using it without knowing users' understanding level is difficult.
Thus, to enhance student comprehension, personalized teaching is essential. 
\newcite{xie2019trends} review personalization research trends from 2007 to 2017, identifying key areas such as the integrating learner preferences, and analyzing individual learning data \cite{chen2008personalized, hwang2010heuristic}.
LLMs can cover them by reinforcement learning from human-feedback (RLHF) that can consider human preferences~\cite{ouyang2022training}.
Current research expects LLMs to present a solution by generating personalized problems and lecture content aligned with learners' comprehension levels \cite{baskara2023exploring}.

However, such instruction-tuning or RLHF-based approaches require task and domain-specific prompts and datasets to train LLMs, especially in the case of targeting personalization. Therefore, considering the various and wide range of fields in education, task-solving by zero-shot approaches is desirable. To achieve that, LLMs need to implicitly adjust text difficulty between user input and its corresponding generated text from LLMs.

For this purpose, our work investigates how LLMs can implicitly adjust text difficulty between user input and its  generated text.
Figure \ref{fig:overview} shows the overview of our investigation that considers the correlation of text difficulties between user input and its generated text by LLMs. To run the experiment, we created a Stack-Overflow dataset by extracting the conversation of questioners and answerers from Stack-Overflow. In addition, to know the adjustment ability of LLMs on conversational text, we also chose the TSCC dataset \cite{caines-etal-2020-teacher} that covers teacher and student conversations.

Experimental results on our Stack-Overflow dataset and the TSCC dataset show that LLMs can handle text difficulty between user input and its generated text in zero-shot learning. Furthermore, we observed that LLMs sometimes surpass the human ability in handling the text difficulty and the importance of instruction-tuning.

\begin{figure*}[t]
  \centering
  \includegraphics[width=\textwidth]{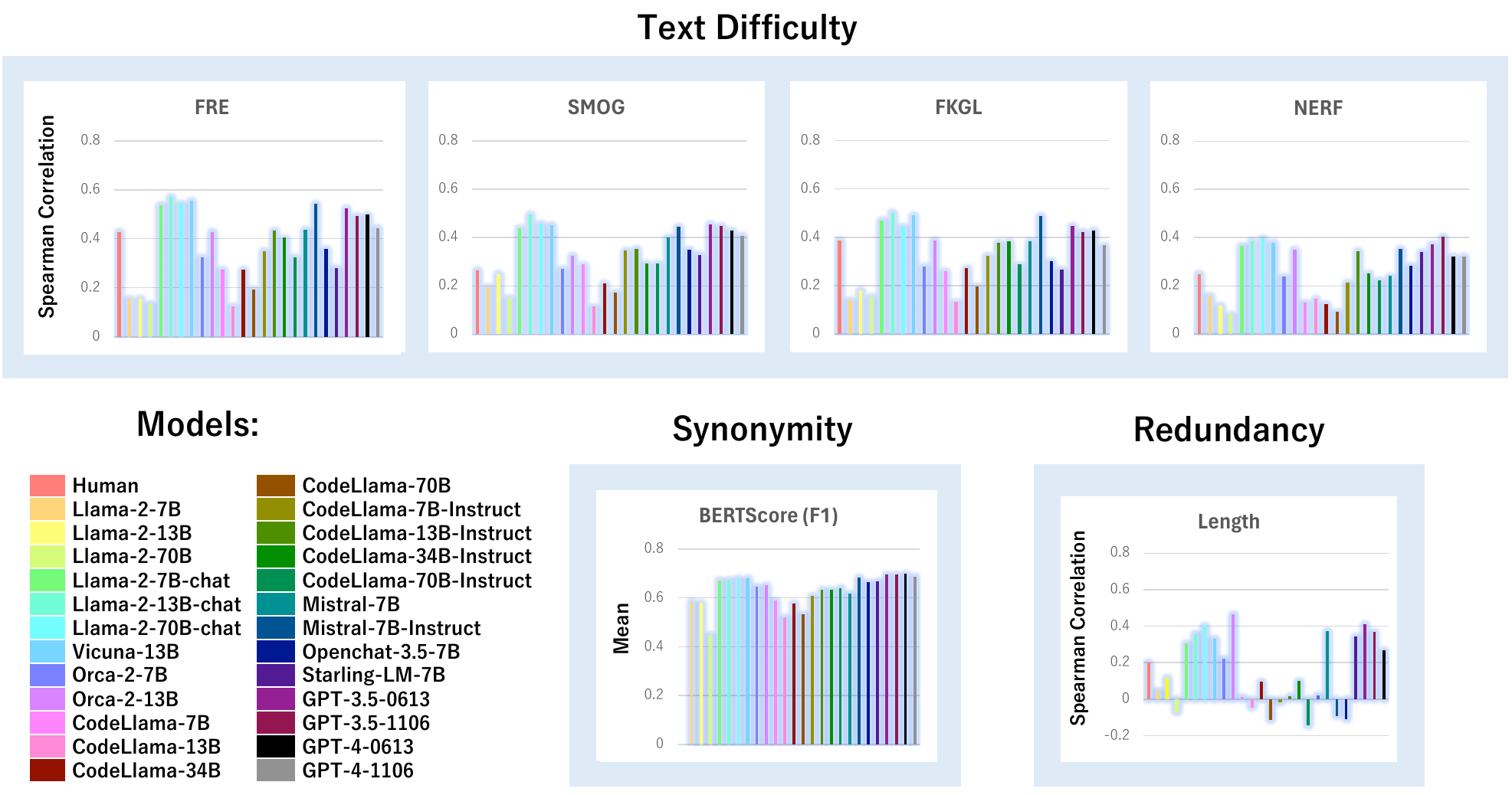}
  \caption{Results on the Stack-Overflow dataset. Note that Table \ref{tab:stack_overflow_normal} and \ref{tab:stack_overflow_normal_mean} in Appendix include the detailed values.}
  \label{fig:stackoverflow_normal}
\end{figure*}

\section{Experimental Setup}
\subsection{Dataset}
We conducted evaluation experiments on two types of datasets: a stack-overflow dataset collected from question and answer sessions, and the Teacher-Student Chatroom Corpus (TSCC) \cite{caines-etal-2020-teacher}, which consists of dialog histories collected during class sessions.

\paragraph{Stack-Overflow}
We created the Stack-Overflow\footnote{\url{https://stackoverflow.com/}} dataset\footnote{See Appendix \ref{sec:appendix:dataset_construction} for the details.}, which consists of 1,000 entries mainly related to programmers' source code and execution environments. It was constructed by scraping question datasets as of July 1, 2023, and extracting question and answer sessions.

\paragraph{TSCC}
\newcite{caines-etal-2020-teacher} published the TSCC, a dataset comprising 260 entries of chat logs between teachers and students collected during class sessions. We extracted dialog histories from the beginning, prefixed them with the labels 'teacher' and 'student', and utilized the dialogues up to just before the first response by the teacher after the 11th turn as input.

\subsection{Models}

To assess the ability of LLMs to adjust text difficulties for users, we compared various models: ChatGPT \cite{ouyang2022training}; LLaMa-2 and LLaMa-2-chat \cite{touvron2023llama2}; Vicuna \cite{zheng2023judging}; CodeLLaMa and CodeLLaMa-Instruct \cite{roziere2023code}; Mistral and Mistral-Instruct \cite{jiang2023mistral}; Orca \cite{mitra2023orca}; OpenChat \cite{wang2023openchat}.\footnote{See Appendix \ref{sec:models_description} for further details.}



Base models are LLaMA-2, CodeLLaMa, and Mistral and instruction-tuned models are LLaMa-2-chat, CodeLLaMa-Instruct, Vicuna, Orca, and OpenChat. Furthermore, to understand the difficulty of this task, we also chose popularly used ChatGPT (GPT-3.5-0613, GPT-3.5-1106, GPT-4-0613, and GPT-4-1106) in our experiment.

To ensure reproducibility, we fixed the random seed and utilized Greedy Search for sentence generation. We detailed inference setting like the total number of input tokens and the maximum number of generation tokens in Appendix \ref{sec:appendix:inference}. 





\subsection{Prompts}
When prompts explicitly indicate the difficulty level, as \newcite{rooein2023know} report, there's a risk of locking in the direction of difficulty adjustment, which might lead to inappropriate personalization not aligned with the user's understanding. Therefore, to evaluate the LLM's ability to adjust difficulty implicitly, we excluded the user's text comprehension level from the prompts or inputs, as detailed in Tables \ref{tab:prompts} and \ref{tab:examples_dialogues} of Appendix \ref{appendix:sec:prompts}.

To assess the effectiveness of prompts, we collected and compared examples of language model outputs across three settings—simple, normal, and complex—within the Stack Overflow dataset, and another setting within the TSCC dataset. Due to the limited space, we only report the result by the normal setting in the main paper. You can see the detailed comparison of the different three prompts in Appendix \ref{sec:detailed_results}.

\subsection{Metrics}
We examine the difficulty adjustment ability of LLMs using three evaluation indicators: text difficulty; synonymity; and text redundancy. In text difficulty and text redundancy, we calculated Spearman's rank correlation coefficient between the scores of input and generated texts. Additionally, we recorded the number of inappropriate text generations (skip rows), such as blanks. Furthermore, we computed the Mean Absolute Error (MAE) and Mean, as detailed in Appendix \ref{sec:detailed_results}.

\paragraph{Text Difficulty} In contexts like language education,  it's crucial for teachers to adapt explanations to match students' vocabulary and comprehension levels. Thus, we consider this ability as a measure of text difficulty. The indicators include traditional ones like FKGL \cite{klare1974assessing}, FRE \cite{kincaid1975derivation}, and SMOG \cite{mc1969smog}, as well as NERF \cite{lee2023traditional}. NERF uses manually created features based on vocabulary difficulty, sentence structure complexity, the diversity of unique words, and bias to formalize text difficulty,offers a more accurate estimation of text difficulty than traditional metrics like FKGL and SMOG.

\paragraph{Synonymity} To assess synonymity, it's essential to determine if LLMs deliver the correct content. Thus, we calculated BERTScore \cite{Zhang2020BERTScore} for texts generated by LLMs, using the collected dataset's texts as references, to ensure LLMs align with the user's intended content.

\paragraph{Redundancy} In contexts like question-answering and education, it's preferable for explanations to be concise and without redundancy. Thus, we investigated if LLMs can produce responses of appropriate length—neither too long nor too short—by comparing the length of LLM-generated texts to the input texts.

\begin{figure*}[t]
  \centering
  \includegraphics[width=\textwidth]{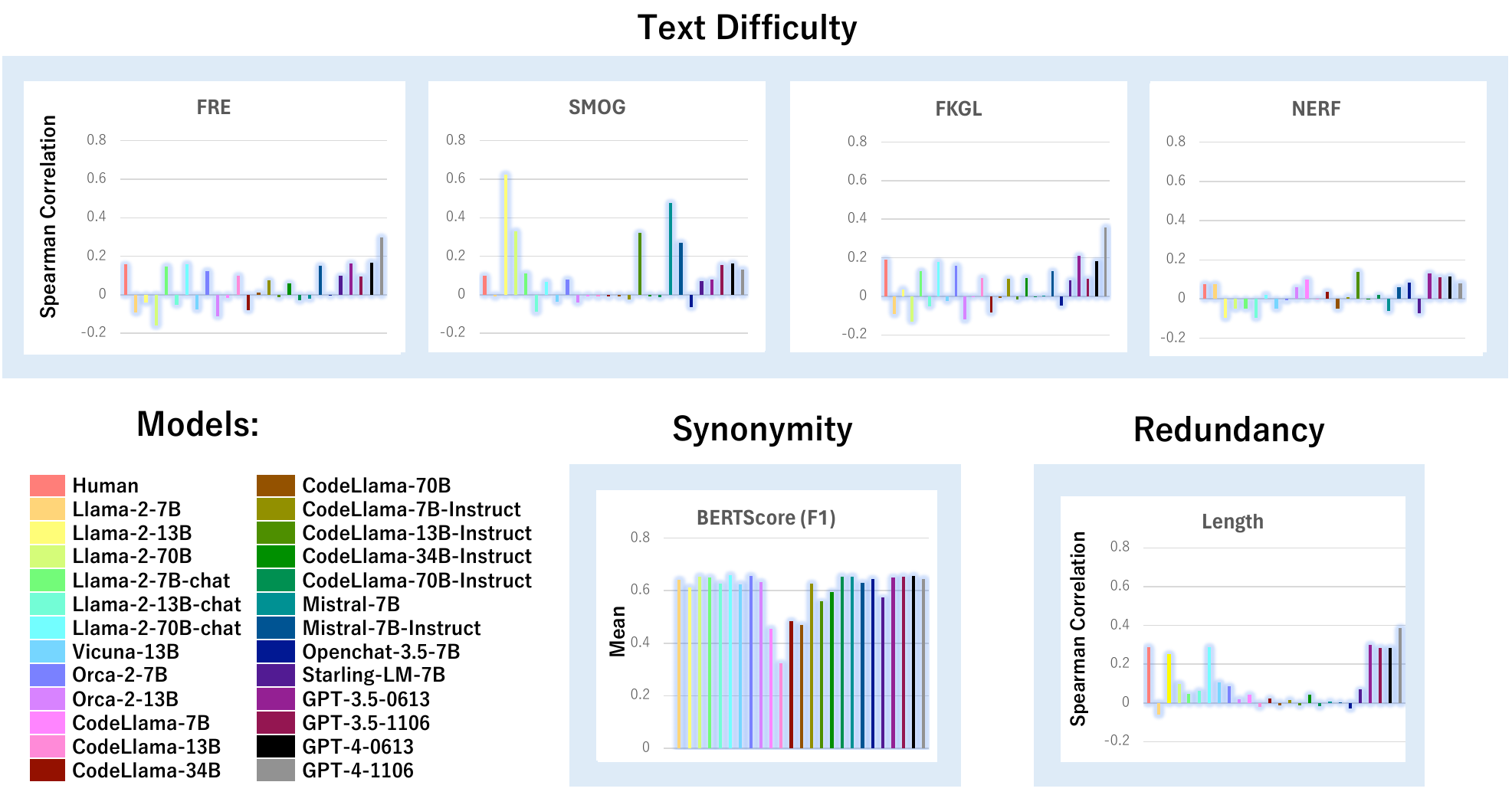}
  \caption{Results on the TSCC dataset. Note that Table \ref{tab:tscc} and \ref{tab:tscc_mean} in Appendix include the detailed values. }
  \label{fig:tscc_results}
\end{figure*}

\section{Results and Discussion} 
\paragraph{Stack-Overflow} Figure \ref{fig:stackoverflow_normal} shows the result on the Stack-Overflow dataset. Although many models score high on BERTScores, the LLaMA-2 base model presents lower scores due to over- and under-generation. This result contrasts LLaMa-2-chat, showing instruction-tuning's effectiveness in considering human responses. Also, LLaMa-2-chat shows great performances in the correlation of text difficulty with other instruction-tuned models, Vicuna-13B and Mistral-7B-Instruct. From the result, we can understand the importance of instruction-tuning in the correlation.

On the other hand, CodeLlama-Instruct, which is instruction-tuned for code generation, shows low performance. Based on the successful result by LLaMa-2-chat, also instruction-tuned from the same model, LLaMa-2, this result indicates the importance of target tasks in instruction-tuning rather than instruction-tuning itself. We can observe a similar relationship between Mistral-7B-Instruct and its instruction-tuned variants, Openchat-3.5-7B and Starling-LM-7B.

Orca shows high performance as an instruction-tuned model. When comparing Orca-2-7B and Orca-2-13B, the findings indicate that Orca-2-13B performs better across all metrics, underscoring the model's adherence to the scaling law. Nevertheless, LLaMA-2-chat maintains strong performance regardless of an increase in model size. Therefore, we can conclude the importance of the instruction-tuning method rather than model parameter size.

LLaMA-2-chat scores comparable to GPT-3.5 and GPT-4 in all metrics. This result is consistent with the human evaluations for helpfulness by LLaMA-2-chat reported in \cite{touvron2023llama2} and shows the potential of open-source models. 

\paragraph{TSCC} Figure~\ref{fig:tscc_results} shows the results of the TSCC dataset. Basically, the correlation coefficient scores for the difficulty of input and generated text are lower than that in the Stack-Overflow dataset, in contrast to the scores in BERTScore. Even in the challenging results by models, we can observe the positive correlations by humans that indicate the validity of this dataset.

In the open-source models, only Llama-2-70B-chat and Mistral-7B-Instruct achieve positive correlations in all metrics, whereas other models sometimes show negative correlations. However, these scores are lower than that of humans, different from the Stack-Overflow dataset. Since the text in the TSCC dataset is often shorter than that in the Stack-Overflow dataset and uses dialogue-specific slang, models need to cover various domains and capture the implicit context of the conversation. Therefore, this result shows room for improvement in the instruction-tuning of open-source models by covering more various domains and diversified conversational text. Furthermore, the inconsistent tendencies of model parameter size support the conclusion induced by the results on the Stack-Overflow dataset that instruction-tuning is more important than the model parameter size.

Regarding GPT-3.5 and GPT-4, the results are remarkably high. These models achieve positive correlations in all metrics similar to humans. Because the details of GPT-3.5 and GPT-4 are not publicly available, we cannot judge what causes this remarkable performance. At least this result indicates the potential of LLMs in handling the correlation of text difficulty between user input and its corresponding response.

\section{Conclusion}
We explored LLMs' ability to implicitly handle text difficulty between user input and generated text by comparing open-source LLMs and ChatGPT models in the Stack-Overflow dataset, based on question answering, and the TSCC dataset, based on dialogue scenarios.

Experimental results on the Stack-Overflow show strong correlations in the text difficulty between texts from LLMs such as LLaMA-2-chat, Vicuna, GPT-3.5, and GPT-4 and their inputs. Notably, sometimes, LLMs even show higher correlation coefficients than human responses, underlining their potential for effective difficulty adjustment in question answering. Furthermore, the experimental results on the TSCC dataset show the difficulty of handling text difficulty between user input and generated text. 

Based on the results, we conclude the importance of instruction-tuning rather that the size of model parameters for implicitly handling text difficulty between user input and generated text by LLMs.

As our future work, we plan to identify preferences that improve this difficulty adjustment ability by examining how LLMs acquire this skill from training data like dialogue histories. 

\newpage
\section{Limitations}
We conducted comparative experiments across various model types, yet we recognize the need for further exploration into datasets and evaluation methodologies.

\paragraph{Datasets}
We chose the Stack-Overflow dataset and TSCC. These datasets focus on distinct domains: coding question-and-answer sessions and dialogue generation for educational guidance, respectively. To effectively evaluate the ability of LLMs to adjust difficulty implicitly, we suggest expanding the evaluations to include a wider variety of domains. This expansion should encompass specialized areas such as law or mathematics and general knowledge domains. Nonetheless, it's crucial to gather responses that are long enough to accurately assess the difficulty of texts produced by LLMs.

\paragraph{Evaluation}
To assess text difficulty, we selected an evaluation metric designed specifically for the English language. Therefore, adapting this evaluation method to other languages requires the use of metrics tailored to each respective language. Additionally, it's vital to verify if the difficulty level of texts produced by LLMs matches users' actual comprehension levels. Although we confirmed that texts generated by models can address certain issues within specific datasets, the extent of the data's contribution to solving problems and the reasons for failures when solutions are not achieved remain unclear.

\section{Ethics Statement}
The LLMs we used in our experiments might contain biases in the datasets utilized during training and the criteria used to ensure their quality. Additionally, the Stack-Overflow dataset employed in this study was collected by the authors themselves. However, for models released after the dataset was collected, there is a possibility that they were trained using the collected dataset.


\bibliography{custom}

\clearpage
\appendix

\section{Inference}
\label{sec:appendix:inference}
\subsection{Hyperparameters}
We conducted 4-bit quantization for inference with a maximum input length of 2048 tokens and a maximum output length of 3072 tokens. We limited the process to a single run since we used the already trained publicly available models in HuggingFace Transformers\footnote{\url{https://huggingface.co}}. We set the random number seed to 42.

\subsection{Prompt}
\label{appendix:sec:prompts}
We analyze LLMs' ability to adjust text difficulty by creating several prompts (see Table \ref{tab:prompts}).

\subsection{Handling Long Inputs}
\label{sec:appendix:long_input}
\begin{figure}
  \centering
  \includegraphics[width=\columnwidth]{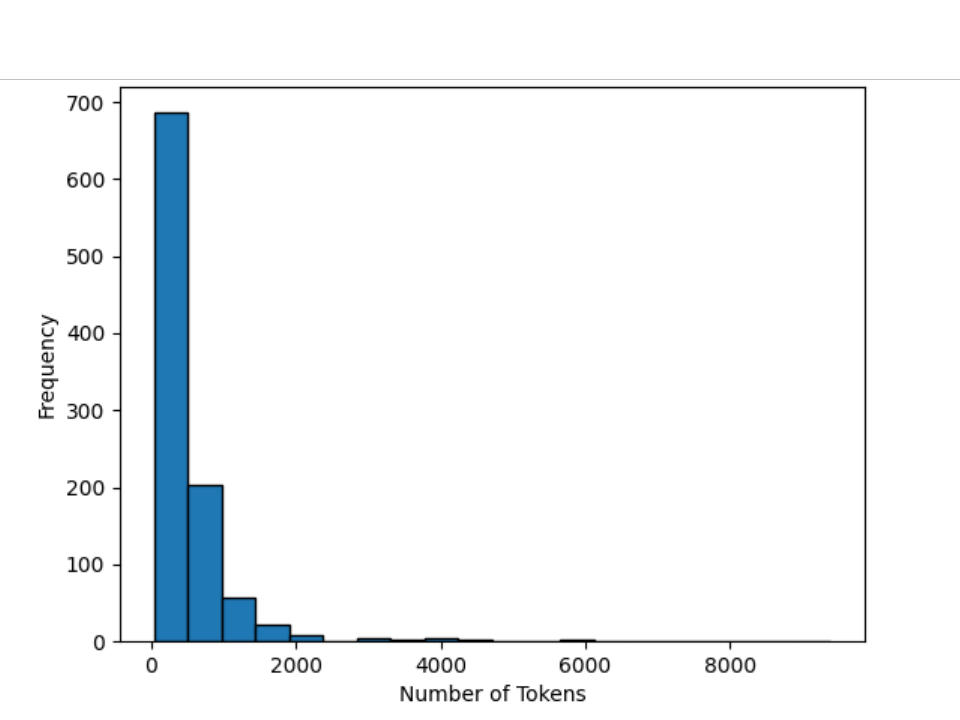}
    \caption{Histgrams of input tokens (Stack-Overflow)}
 \label{fig:token_hist}
\end{figure}

Figure \ref{fig:token_hist} shows a histogram of the number of tokens calculated using the tokenizer of Llama-2-7B~\cite{touvron2023llama} for the input data of the Stack-Overflow dataset. In Figure \ref{fig:token_hist}, 97.0\% of all input data has 2048 tokens or fewer, 98.1\% has 3072 tokens or fewer, and 1.9\% has more than 3072 tokens. To evaluate whether the model has acquired the ability to adjust difficulty levels in the outputs it generates for input sentences, it is not necessary to consider all input sentences; it is considered possible to capture the content of many input sentences sufficiently with 2048 tokens. Therefore, to standardize the length of input and output sentences generated, the input to the model was truncated to up to 2048 tokens, and the maximum number of tokens generated was adjusted to match the input tokens, resulting in 3072 tokens.

\subsection{Total Computational Budget}
We utilized GPUs for a total of 2,500 hours to generate texts. Additionally, we incurred costs of \$246.36 through the OpenAI API for inference.

\begin{table*}[ht]
\centering
\small
\begin{tabular}{ll}
\toprule
\multicolumn{2}{c}{\textbf{Stack-Overflow}} \\
\midrule
Setting   & Prompt \\
\midrule
    Normal & 
         \#\#\# Question : \{\textcolor{blue}{$Title$}\} \{\textcolor{red}{$Question$}\}\\
    \midrule
    Simple & 

         Please respond to the question using simple and user-friendly language.\\
         & \#\#\# Question : \{\textcolor{blue}{$Title$}\} \{\textcolor{red}{$Question$}\}\\ 
    \midrule
    Complex & Please respond to the question using complex and less user-friendly language.\\
         & \#\#\# Question : \{\textcolor{blue}{$Title$}\} \{\textcolor{red}{$Question$}\}\\ 
\bottomrule
\toprule
\multicolumn{2}{c}{\textbf{TSCC}} \\
\midrule
         \multicolumn{2}{l}{Please generate a response from the teacher to the student in the ongoing dialogue.} \\
         \multicolumn{2}{l}{\#\#\# Dialogue :  \{\textcolor{red}{$Dialogue$}\}}\\ 
\bottomrule
\end{tabular}
\caption{Prompts for each setting. Note that TSCC has only one prompt.}
\label{tab:prompts}
\end{table*}

\begin{table*}[t]
\centering
\small
\begin{tabular}{ll}
\toprule
\multirow{13}{*}{Input} & Please generate a response from the teacher to the student in the ongoing dialogue. \\
& \#\#\# Dialogue:student: Hi! \\
& teacher: Hi $<$STUDENT$>$!\\
& teacher: Everything alright with the chatroom for you?\\
& student: I tried to use it a few seconds ago and I couldn't change my name, but now it is working, thanks.\\
& student: How are you?\\
& teacher: Oh good!\\
& teacher: Fine, thank you! It's summer here at last, we've had a week of non-stop sunshine!\\
& teacher: How are you?\\
& student: I'm fine thank you! It looks like summer has arrived here too!\\
& student: Even though we still had a couple of storms...\\
& student: with hail and everything\\
& teacher:\\
\midrule
Output & Ooh, I hope you're not too badly affected by them!\\
\bottomrule
\end{tabular}
\caption{Examples of dialogues (Starling-7B)}
\label{tab:examples_dialogues}
\end{table*}

\section{Detailed Results}
 We calculate the scores using pairs of input texts and their generated texts (human responses). Additionally, we calculate document length based on the number of characters.
\label{sec:detailed_results}
\subsection{Spearman Correlation}
We compare LLMs' ability to adjust text difficulty and redundancy using the Spearman correlation. Tables \ref{tab:stack_overflow_normal}--\ref{tab:tscc} show the actual scores.

\label{sec:spearman_corr}
\begin{table*}[ht]
\centering
\small
\begin{tabular}{lccccr}
\toprule
Models & FRE & SMOG & FKGL & NERF  & Length \\
\midrule
Human & 0.428 & 0.265 & 0.387 & 0.248 & 0.203 \\
\midrule
Llama-2-7B & 0.157 & 0.196 & 0.140 & 0.159 & 0.047  \\
Llama-2-13B & 0.157 & 0.249 & 0.182 & 0.118 & 0.119 \\
Llama-2-70B & 0.133 & 0.150 & 0.154 & 0.082 & -0.070  \\
\midrule
Llama-2-7B-chat & 0.538 & 0.438 & 0.469 & 0.364  & 0.306  \\
Llama-2-13B-chat & 0.571 & 0.495 & 0.502 & 0.386  & 0.356 \\
Llama-2-70B-chat & 0.545 & 0.459 & 0.445 & 0.397  & 0.402  \\
Vicuna-13B & 0.555 & 0.452 & 0.491 & 0.380  & 0.333  \\
Orca-2-7B & 0.324 & 0.271 & 0.280 & 0.239  & 0.226  \\
Orca-2-13B & 0.426 & 0.325 & 0.388 & 0.350  & 0.467  \\
\midrule
CodeLlama-7B & 0.275 & 0.288 & 0.260 & 0.130  & 0.016  \\
CodeLlama-13B & 0.123 & 0.114 & 0.135 & 0.149  & -0.043  \\
CodeLlama-34B & 0.275 & 0.212 & 0.275 & 0.125  & 0.098  \\
CodeLlama-70B & 0.192 & 0.173 & 0.199 & 0.093  & -0.113  \\
CodeLlama-7B-Instruct & 0.349 & 0.347 & 0.325 & 0.215  & -0.018  \\
CodeLlama-13B-Instruct & 0.433 & 0.354 & 0.376 & 0.343 & 0.017  \\
CodeLlama-34B-Instruct & 0.405 & 0.294 & 0.383 & 0.251 & 0.102 \\
CodeLlama-70B-Instruct & 0.322 & 0.293 & 0.288 & 0.222  & -0.143 \\
\midrule
Mistral-7B & 0.361 & 0.343 & 0.316 & 0.260  & 0.042  \\
Mistral-7B-Instruct & 0.542 & 0.443 & 0.489 & 0.353  & 0.375  \\
Openchat-3.5-7B & 0.359 & 0.348 & 0.300 & 0.283  & -0.092  \\
Starling-LM-7B & 0.281 & 0.328 & 0.265 & 0.340  & -0.110  \\
\midrule
GPT-3.5-0613 & 0.523 & 0.455 & 0.448 & 0.373  & 0.342  \\
GPT-3.5-1106 & 0.492 & 0.448 & 0.422 & 0.405  & 0.414  \\
GPT-4-0613 & 0.498 & 0.430 & 0.428 & 0.323  & 0.370  \\
GPT-4-1106 & 0.443 & 0.407 & 0.366 & 0.322  & 0.268  \\
\bottomrule
\end{tabular}
\caption{Stack-Overflow Normal Setting (Spearman Correlation)}
\label{tab:stack_overflow_normal}
\end{table*}

\begin{table*}[ht]
\centering
\small
\begin{tabular}{lccccr}
\toprule
Models & \begin{tabular}{c} FRE \end{tabular} & SMOG & FKGL & NERF  & Length \\
\midrule
Human & 0.428 & 0.265 & 0.387 & 0.248 & 0.203  \\
\midrule
Llama-2-7B & 0.128 & 0.222 & 0.117 & 0.109  & 0.070  \\
Llama-2-13B & 0.143 & 0.221 & 0.118 & 0.170  & 0.019  \\
Llama-2-70B & 0.085 & 0.142 & 0.100 & 0.053  & -0.129  \\
\midrule
Llama-2-7B-chat & 0.541 & 0.492 & 0.483 & 0.331  & 0.395 \\
Llama-2-13B-chat & 0.562 & 0.490 & 0.472 & 0.331  & 0.357  \\
Llama-2-70B-chat & 0.560 & 0.500 & 0.492 & 0.391  & 0.454  \\
Vicuna-13B & 0.503 & 0.428 & 0.460 & 0.351  & 0.335\\
Orca-2-7B & 0.238 & 0.139 & 0.195 & 0.164  & 0.186 \\
Orca-2-13B & 0.322 & 0.289 & 0.308 & 0.300  & 0.400 \\
\midrule
CodeLlama-7B & 0.332 & 0.341 & 0.316 & 0.215  & -0.054 \\
CodeLlama-13B & 0.182 & 0.238 & 0.200 & 0.140  & -0.057  \\
CodeLlama-34B & 0.154 & 0.167 & 0.133 & 0.081  & 0.104 \\
CodeLlama-70B & 0.075 & 0.120 & 0.128 & 0.053& -0.067 \\
CodeLlama-7B-Instruct & 0.460 & 0.392 & 0.412 & 0.338 & -0.076  \\
CodeLlama-13B-Instruct & 0.362 & 0.343 & 0.307 & 0.289 & -0.078  \\
CodeLlama-34B-Instruct & 0.435 & 0.370 & 0.369 & 0.265  & 0.265  \\
CodeLlama-70B-Instruct & 0.306 & 0.171 & 0.230 & 0.313 & -0.379  \\
\midrule
Mistral-7B & 0.461 & 0.400 & 0.418 & 0.257  & 0.040 \\
Mistral-7B-Instruct & 0.530 & 0.495 & 0.480 & 0.338  & 0.481  \\
Openchat-3.5-7B & 0.424 & 0.369 & 0.369 & 0.280 & 0.130  \\
Starling-LM-7B & 0.279 & 0.312 & 0.259 & 0.329  & -0.149 \\
\midrule
GPT-3.5-0613 & 0.503 & 0.456 & 0.430 & 0.368  & 0.430 \\
GPT-3.5-1106 & 0.472 & 0.442 & 0.401 & 0.367  & 0.496 \\
GPT-4-0613 & 0.413 & 0.417 & 0.350 & 0.269  & 0.461 \\
GPT-4-1106 & 0.432 & 0.397 & 0.363 & 0.323 & 0.335  \\
\bottomrule
\end{tabular}
\caption{Stack-Overflow Simple Setting (Spearman Correlation)}
\label{tab:stack_oveflow_simple}
\end{table*}

\begin{table*}[hp]
\centering
\small
\begin{tabular}{lccccr}
\toprule
Models & FRE & SMOG & FKGL & NERF  & Length  \\
\midrule
Human & 0.428 & 0.265 & 0.387 & 0.248 & 0.203  \\
\midrule
Llama-2-7B & 0.107 & 0.221 & 0.105 & 0.092  & 0.070  \\
Llama-2-13B & 0.165 & 0.221 & 0.142 & 0.213  & 0.042\\
Llama-2-70B & 0.049 & 0.137 & 0.064 & 0.038  & -0.130  \\
\midrule
Llama-2-7B-chat & 0.487 & 0.397 & 0.388 & 0.216 & 0.144  \\
Llama-2-13B-chat & 0.542 & 0.467 & 0.464 & 0.342  & 0.218  \\
Llama-2-70B-chat & 0.535 & 0.461 & 0.463 & 0.298  & 0.319 \\
Vicuna-13B & 0.458 & 0.352 & 0.390 & 0.285  & 0.273  \\
Orca-2-7B & 0.224 & 0.141 & 0.181 & 0.158  & 0.108  \\
Orca-2-13B & 0.296 & 0.271 & 0.264 & 0.229  & 0.285  \\
\midrule
CodeLlama-7B & 0.346 & 0.313 & 0.315 & 0.233 & -0.025  \\
CodeLlama-13B & 0.143 & 0.241 & 0.174 & 0.108  & 0.000  \\
CodeLlama-34B & 0.084 & 0.134 & 0.099 & -0.011  & 0.134  \\
CodeLlama-70B & 0.089 & 0.182 & 0.144 & 0.058  & -0.087  \\
CodeLlama-7B-Instruct & 0.440 & 0.359 & 0.389 & 0.321 & -0.077  \\
CodeLlama-13B-Instruct & 0.288 & 0.280 & 0.270 & 0.212  & -0.105  \\
CodeLlama-34B-Instruct & 0.471 & 0.409 & 0.425 & 0.236 & 0.272  \\
CodeLlama-70B-Instruct & 0.333 & 0.257 & 0.294 & 0.253  & -0.169 \\
\midrule
Mistral-7B & 0.438 & 0.400 & 0.384 & 0.240  & 0.023\\
Mistral-7B-Instruct & 0.431 & 0.434 & 0.389 & 0.287 & 0.430  \\
Openchat-3.5-7B & 0.511 & 0.415 & 0.432 & 0.343& 0.191 \\
Starling-LM-7B & 0.305 & 0.255 & 0.274 & 0.295  & -0.218 \\
\midrule
GPT-3.5-0613 & 0.404 & 0.340 & 0.341 & 0.284 & 0.374 \\
GPT-3.5-1106 & 0.276 & 0.266 & 0.231 & 0.118 & 0.475 \\
GPT-4-0613 & 0.297 & 0.274 & 0.230 & 0.174  & 0.513  \\
GPT-4-1106 & 0.370 & 0.304 & 0.311 & 0.197 & 0.297 \\
\bottomrule
\end{tabular}
\caption{Stack-Overflow Complex Setting (Spearman Correlation)}
\label{tab:stack_overflow_complex}
\end{table*}

\begin{table*}[hp]
\centering
\small
\begin{tabular}{lrrrrr}
\toprule
Models & \begin{tabular}{c} FRE \end{tabular} &  SMOG & FKGL & NERF & Length  \\
\midrule
Human & 0.157 & 0.098 & 0.192 & 0.075 & 0.288 \\
\midrule
Llama-2-7B & -0.093 & -0.010 & -0.094 & 0.075  & -0.062 \\
Llama-2-13B & -0.041 & 0.622 & 0.035 & -0.097  & 0.252 \\
Llama-2-70B & -0.162 & 0.329 & -0.129 & -0.049  & 0.100 \\
\midrule
Llama-2-7B-chat & 0.146 & 0.111 & 0.131 & -0.048  & 0.047 \\
Llama-2-13B-chat & -0.052 & -0.089 & -0.051 & -0.095  & 0.061  \\
Llama-2-70B-chat & 0.159 & 0.066 & 0.178 & 0.022  & 0.288  \\
Vicuna-13B & -0.076 & -0.037 & -0.024 & -0.049 & 0.104 \\
Orca-2-7B & 0.124 & 0.079 & 0.160 & -0.007 & 0.087  \\
Orca-2-13B & -0.111 & -0.041 & -0.120 & 0.058  & 0.021  \\
\midrule
CodeLlama-7B & -0.016 & -0.010 & -0.001 & 0.099 & 0.044  \\
CodeLlama-13B & 0.098 & -0.010 & 0.096 & 0.002  & -0.020  \\
CodeLlama-34B & -0.082 & -0.010 & -0.083 & 0.037  & 0.024 \\
CodeLlama-70B & 0.013 & -0.010 & -0.006 & -0.049  & -0.013  \\
CodeLlama-7B-Instruct & 0.074 & -0.024 & 0.093 & 0.008 & 0.014 \\
CodeLlama-13B-Instruct & -0.013 & 0.321 & -0.016 & 0.141  & -0.012 \\
CodeLlama-34B-Instruct & 0.062 & -0.010 & 0.096 & -0.002 & 0.044 \\
CodeLlama-70B-Instruct & -0.029 & -0.013 & -0.003 & 0.019  & -0.017 \\
\midrule
Mistral-7B & -0.022 & 0.478 & 0.002 & -0.061  & 0.007 \\
Mistral-7B-Instruct & 0.149 & 0.270 & 0.130 & 0.059  & 0.001  \\
Openchat-3.5-7B & -0.007 & -0.065 & -0.049 & 0.084  & -0.031 \\
Starling-LM-7B & 0.096 & 0.071 & 0.084 & -0.071  & 0.069 \\
\midrule
GPT-3.5-0613 & 0.163 & 0.076 & 0.210 & 0.130 & 0.301 \\
GPT-3.5-1106 & 0.095 & 0.152 & 0.091 & 0.110  & 0.285 \\
GPT-4-0613 & 0.167 & 0.163 & 0.184 & 0.113  & 0.285  \\
GPT-4-1106 & 0.300 & 0.132 & 0.357 & 0.080  & 0.388  \\

\bottomrule
\end{tabular}
\caption{TSCC Setting (Spearman Correlation)}
\label{tab:tscc}
\end{table*}

\subsection{Mean Scores}
In Table \ref{tab:stack_overflow_normal_mean}--\ref{tab:tscc_mean} , we observe that models, with the exception of CodeLLaMa, which have enhanced ability to adjust difficulty, tend to produce shorter texts. This indicates that instruction-tuning likely facilitates the development of skills to appropriately regulate response lengths. Although this study evaluated the length of texts generated by LLMs in comparison to their original lengths, the ideal text length should naturally vary from one user to another. Thus, aside from extreme cases like CodeLLaMa, there's a need to explore effective evaluation methods for determining the suitable length of LLM-generated texts and to establish credible criteria for assessing longer text outputs.Additionally, GPT-4-1106 produced longer texts than those by previous versions, GPT-3.5 and GPT-4, suggesting it might use longer sequences for training. This indicates that GPT-4 may generate redundant responses without specific tuning prompts.

\label{sec:mean}
\begin{table*}[hp]
\centering
\small
\begin{tabular}{lrrrrcc}
\toprule
Models & \begin{tabular}{c} FRE \end{tabular} & SMOG & FKGL & NERF & BERTScore (F1) &\begin{tabular}{c} Length \end{tabular}\\
\midrule
Human & 42.358 & 11.228 & 11.557 & 6.765 & -- & 1729.109 \\
\midrule
Llama-2-7B & -3.915 & 8.785 & 21.369 & 3.843 & 0.587 & 5745.329 \\
Llama-2-13B & -169.850 & 6.917 & 49.335 & 30.439 & 0.581 & 4583.894 \\
Llama-2-70B & 64.929 & 6.606 & 9.636 & 3.617 & 0.448 & 3995.069 \\
\midrule
Llama-2-7B-chat & 49.029 & 11.994 & 11.040 & 3.758 & 0.672 & 1894.843 \\
Llama-2-13B-chat & 0.272 & 11.769 & 17.712 & 3.827 & 0.673 & 2100.051 \\
Llama-2-70B-chat & 49.231 & 12.006 & 11.013 & 4.200 & 0.679 & 1965.053 \\
Vicuna-13B & 48.784 & 11.442 & 10.807 & 4.627 & 0.682 & 1592.608 \\
Orca-2-7B & 74.026 & 8.663 & 6.453 & 2.839 & 0.646 & 1164.153 \\
Orca-2-13B & 72.520 & 8.637 & 6.708 & 3.072 & 0.652 & 1213.115 \\
\midrule
CodeLlama-7B & 19.119 & 9.329 & 19.519 & 10.321 & 0.591 & 5979.621 \\
CodeLlama-13B & -3.200 & 8.839 & 20.220 & 5.913 & 0.520 & 5309.517 \\
CodeLlama-34B & 34.064 & 7.996 & 13.963 & 3.287 & 0.577 & 3680.992 \\
CodeLlama-70B & 13.301 & 8.062 & 19.851 & 4.179 & 0.534 & 5884.443 \\
CodeLlama-7B-Instruct & 39.036 & 10.038 & 13.560 & 2.580 & 0.609 & 5778.107 \\
CodeLlama-13B-Instruct & 33.698 & 10.659 & 13.536 & 2.181 & 0.633 & 5014.724 \\
CodeLlama-34B-Instruct & 38.577 & 9.585 & 12.440 & 3.540 & 0.635 & 3912.342 \\
CodeLlama-70B-Instruct & 33.505 & 10.033 & 14.082 & 3.550 & 0.640 & 5985.720 \\
\midrule
Mistral-7B & 30.479 & 10.171 & 16.333 & 1.278 & 0.619 & 5014.421 \\
Mistral-7B-Instruct & 43.342 & 11.579 & 12.014 & 4.425 & 0.683 & 1901.848 \\
Openchat-3.5-7B & 33.378 & 10.829 & 12.943 & 2.333 & 0.664 & 5747.161 \\
Starling-LM-7B & 8.850 & 11.288 & 16.642 & 3.150 & 0.670 & 6941.246 \\
\midrule
GPT-3.5-0613 & 47.954 & 11.901 & 10.775 & 4.939 & 0.697 & 1392.241 \\
GPT-3.5-1106 & 47.598 & 12.308 & 11.199 & 5.157 & 0.695 & 1428.607 \\
GPT-4-0613 & 54.886 & 11.190 & 9.617 & 4.348 & 0.699 & 1323.731 \\
GPT-4-1106 & 50.680 & 12.286 & 10.829 & 5.660 & 0.688 & 2328.291 \\
\bottomrule
\end{tabular}
\caption{Stack-Overflow Normal Setting (Mean)}
\label{tab:stack_overflow_normal_mean}
\end{table*}

\begin{table*}[hp]
\centering
\small
\begin{tabular}{lrrrrcr}
\toprule
Models & \begin{tabular}{c} FRE \end{tabular} & SMOG & FKGL & NERF & BERTScore (F1) & \begin{tabular}{c} Length  \end{tabular} \\
\midrule
Human & 42.358 & 11.228 & 11.557 & 6.765 & -- & 1729.109 \\
\midrule
Llama-2-7B & -44.747 & 9.201 & 30.078 & 9.021 & 0.591 & 6144.573 \\
Llama-2-13B & -102.317 & 7.000 & 49.811 & 60.371 & 0.574 & 5583.394 \\
Llama-2-70B & 15.357 & 7.387 & 23.598 & 18.986 & 0.499 & 4715.437 \\
\midrule
Llama-2-7B-chat & 52.186 & 11.782 & 10.514 & 3.732 & 0.672 & 1668.559 \\
Llama-2-13B-chat & 14.154 & 11.539 & 15.701 & 3.878 & 0.673 & 1883.881 \\
Llama-2-70B-chat & 50.721 & 11.805 & 10.640 & 4.183 & 0.680 & 1723.086 \\
Vicuna-13B & 53.171 & 10.886 & 10.121 & 4.274 & 0.681 & 1524.795 \\
Orca-2-7B & 66.388 & 6.515 & 6.583 & 1.268 & 0.609 & 933.419 \\
Orca-2-13B & 92.495 & 7.521 & 3.435 & 1.990 & 0.634 & 1091.195 \\
\midrule
CodeLlama-7B & -49.313 & 9.495 & 28.247 & 5.624 & 0.583 & 6344.100 \\
CodeLlama-13B & 39.978 & 8.331 & 13.847 & 1.397 & 0.525 & 5727.908 \\
CodeLlama-34B & 45.189 & 7.562 & 12.502 & 4.665 & 0.548 & 3846.843 \\
CodeLlama-70B & 22.740 & 7.409 & 18.632 & 3.908 & 0.493 & 5632.746 \\
CodeLlama-7B-Instruct & 21.654 & 10.439 & 15.220 & 1.592 & 0.629 & 6342.817 \\
CodeLlama-13B-Instruct & 48.177 & 9.885 & 10.735 & 1.162 & 0.609 & 5553.601 \\
CodeLlama-34B-Instruct & 53.111 & 10.520 & 9.878 & 3.002 & 0.646 & 2935.139 \\
CodeLlama-70B-Instruct & 39.935 & 11.960 & 12.655 & 2.310 & 0.643 & 8288.849 \\
\midrule
Mistral-7B & 39.831 & 10.262 & 13.967 & 0.920 & 0.624 & 4611.053 \\
Mistral-7B-Instruct & 50.899 & 11.490 & 10.790 & 3.814 & 0.676 & 1647.081 \\
Openchat-3.5-7B & 46.104 & 11.085 & 10.975 & 3.363 & 0.674 & 3931.610 \\
Starling-LM-7B & 19.575 & 11.430 & 13.878 & 3.566 & 0.671 & 7286.648 \\
\midrule
GPT-3.5-0613 & 53.527 & 11.522 & 9.950 & 4.354 & 0.694 & 1181.735 \\
GPT-3.5-1106 & 50.124 & 11.592 & 10.836 & 4.298 & 0.700 & 1009.199 \\
GPT-4-0613 & 59.545 & 10.902 & 8.972 & 3.842 & 0.694 & 1004.923 \\
GPT-4-1106 & 52.309 & 12.131 & 10.700 & 5.333 & 0.688 & 2112.660 \\
\bottomrule
\end{tabular}
\caption{Stack-Overflow Simple Setting (Mean)}
\label{tab:stack_overflow_simple_mean}
\end{table*}

\begin{table*}[hp]
\centering
\small
\begin{tabular}{lrrrrcc}
\toprule
Models & \begin{tabular}{c} FRE \end{tabular} & SMOG & FKGL & NERF & BERTScore (F1) & \begin{tabular}{c} Length  \end{tabular} \\
\midrule
Human & 42.358 & 11.228 & 11.557 & 6.765 & -- & 1729.109 \\
\midrule
Llama-2-7B & -52.236 & 8.987 & 33.757 & 9.682 & 0.589 & 6134.990 \\
Llama-2-13B & -64.823 & 7.199 & 41.513 & 57.876 & 0.578 & 5596.936 \\
Llama-2-70B & 27.943 & 6.778 & 19.205 & 13.451 & 0.453 & 4635.005 \\
\midrule
Llama-2-7B-chat & 49.262 & 12.313 & 11.097 & 4.149 & 0.667 & 2018.452 \\
Llama-2-13B-chat & 44.077 & 11.584 & 11.635 & 3.836 & 0.666 & 2021.876 \\
Llama-2-70B-chat & 46.869 & 12.633 & 11.660 & 4.582 & 0.677 & 1996.049 \\
Vicuna-13B & -153.948 & 11.281 & 39.172 & 4.811 & 0.668 & 1730.558 \\
Orca-2-7B & 102.040 & 7.175 & 1.910 & 1.479 & 0.609 & 1062.560 \\
Orca-2-13B & 78.777 & 9.046 & 5.805 & 2.742 & 0.638 & 1318.739 \\
\midrule
CodeLlama-7B & 8.682 & 9.430 & 20.338 & 6.238 & 0.582 & 6280.695 \\
CodeLlama-13B & 37.556 & 8.202 & 14.556 & 1.852 & 0.512 & 5164.743 \\
CodeLlama-34B & 50.118 & 6.954 & 11.484 & 10.591 & 0.513 & 3610.031 \\
CodeLlama-70B & 23.125 & 7.549 & 18.884 & 3.738 & 0.490 & 5581.595 \\
CodeLlama-7B-Instruct & 45.469 & 10.016 & 12.308 & 1.235 & 0.608 & 6487.346 \\
CodeLlama-13B-Instruct & 63.227 & 9.083 & 8.981 & 1.438 & 0.545 & 5600.361 \\
CodeLlama-34B-Instruct & 59.502 & 10.586 & 8.969 & 2.824 & 0.631 & 2802.091 \\
CodeLlama-70B-Instruct & 57.045 & 10.067 & 9.969 & 1.521 & 0.596 & 7059.423 \\
\midrule
Mistral-7B & 40.518 & 10.209 & 14.164 & 1.431 & 0.618 & 4777.848 \\
Mistral-7B-Instruct & 44.273 & 12.599 & 12.254 & 3.522 & 0.671 & 2033.776 \\
Openchat-3.5-7B & 44.957 & 12.189 & 11.445 & 4.209 & 0.675 & 3517.100 \\
Starling-LM-7B & 30.399 & 12.958 & 13.320 & 3.515 & 0.670 & 8060.675 \\
\midrule
GPT-3.5-0613 & 48.464 & 12.475 & 11.044 & 4.656 & 0.684 & 1380.164 \\
GPT-3.5-1106 & 39.075 & 14.527 & 13.211 & 5.053 & 0.655 & 1233.771 \\
GPT-4-0613 & 44.493 & 13.819 & 12.164 & 5.314 & 0.666 & 1615.374 \\
GPT-4-1106 & 36.727 & 14.807 & 13.683 & 7.223 & 0.674 & 2661.302 \\
\bottomrule
\end{tabular}
\caption{Stack-Overflow Complex Setting (Mean)}
\label{tab:stack_overflow_complex_mean}
\end{table*}

\begin{table*}[hp]
\centering
\small
\begin{tabular}{lrcrrcr}
\toprule
Models & \begin{tabular}{c} FRE \end{tabular} & SMOG & FKGL & NERF & BERTScore (F1) & \begin{tabular}{c} Length  \end{tabular} \\
\midrule
Human & 88.507 & 0.567 & 3.119 & -0.393 & -- & 68.677 \\
\midrule
Llama-2-7B & 82.350 & 0.025 & 5.864 & 0.203 & 0.642 & 113.088 \\
Llama-2-13B & 108.542 & 0.144 & 1.152 & 4.904 & 0.613 & 170.804 \\
Llama-2-70B & 110.196 & 0.125 & -0.733 & 13.679 & 0.653 & 88.888 \\
\midrule
Llama-2-7B-chat & 90.516 & 0.861 & 2.545 & 0.359 & 0.652 & 88.362 \\
Llama-2-13B-chat & 46.364 & 1.755 & 8.728 & 0.826 & 0.628 & 364.665 \\
Llama-2-70B-chat & 91.138 & 1.384 & 2.616 & 6.912 & 0.658 & 131.462 \\
Vicuna-13B & 88.828 & 0.390 & 2.607 & 9.391 & 0.623 & 89.227 \\
Orca-2-7B & 98.840 & 0.326 & 1.311 & -0.221 & 0.655 & 62.408 \\
Orca-2-13B & 76.594 & 0.472 & 4.229 & -0.280 & 0.634 & 84.462 \\
\midrule
CodeLlama-7B & 124.331 & 0.012 & -0.395 & -0.337 & 0.454 & 78.050 \\
CodeLlama-13B & 152.196 & 0.039 & -7.438 & 1.453 & 0.324 & 78.938 \\
CodeLlama-34B & 131.738 & 0.034 & -4.277 & 22.400 & 0.483 & 97.919 \\
CodeLlama-70B & 127.126 & 0.012 & -1.671 & 14.973 & 0.469 & 181.892 \\
CodeLlama-7B-Instruct & 104.029 & 0.141 & 0.207 & -0.557 & 0.626 & 66.923 \\
CodeLlama-13B-Instruct & 107.991 & 0.090 & 1.725 & 7.333 & 0.558 & 117.608 \\
CodeLlama-34B-Instruct & 117.322 & 0.036 & -1.806 & 42.973 & 0.594 & 221.046 \\
CodeLlama-70B-Instruct & 95.991 & 0.062 & 3.783 & 13.962 & 0.652 & 172.177 \\
\midrule
Mistral-7B & 107.466 & 0.056 & 1.375 & 2.323 & 0.652 & 114.004 \\
Mistral-7B-Instruct & 102.654 & 0.100 & 1.192 & 16.965 & 0.629 & 225.177 \\
Openchat-3.5-7B & 95.955 & 1.367 & 1.599 & 3.411 & 0.644 & 531.673 \\
Starling-LM-7B & 66.350 & 7.813 & 7.132 & 1.823 & 0.573 & 5100.092 \\
\midrule
GPT-3.5-0613 & 80.366 & 6.560 & 4.636 & 1.877 & 0.651 & 204.042 \\
GPT-3.5-1106 & 80.508 & 4.976 & 4.528 & 1.715 & 0.652 & 150.992 \\
GPT-4-0613 & 80.493 & 5.217 & 4.444 & 1.775 & 0.656 & 157.319 \\
GPT-4-1106 & 77.535 & 7.843 & 5.283 & 2.417 & 0.643 & 261.388 \\

\bottomrule
\end{tabular}
\caption{TSCC Setting (Mean)}
\label{tab:tscc_mean}
\end{table*}

\subsection{Mean Absolute Error}
Tables \ref{tab:stack_overflow_normal_mae}--\ref{tab:tscc_mae} show that mean absolute error between input texts and generated texts.
As shown in Table \ref{tab:stack_overflow_normal_mae}--\ref{tab:tscc_mae}, we observed the tendency similar to the Spearman correlation.
Additionally, well instruction-tuned models, such as LLaMA-2-chat and GPT4 score low mean absolute error.

\label{sec:mae}
\begin{table*}[hp]
\centering
\small
\begin{tabular}{lrcrrr}
\toprule
Models & \begin{tabular}{c} FRE \end{tabular} & SMOG & FKGL & NERF & \begin{tabular}{c} Length \end{tabular} \\
\midrule
Human & 25.878 & 3.526 & 4.575 & 2.895 & 1243.833 \\
\midrule
Llama-2-7B & 97.339 & 4.577 & 18.853 & 10.719 & 4457.702 \\
Llama-2-13B & 251.946 & 5.414 & 44.864 & 38.081 & 3510.847 \\
Llama-2-70B & 97.945 & 6.168 & 18.376 & 10.124 & 3295.110 \\
\midrule
Llama-2-7B-chat & 19.359 & 2.191 & 3.481 & 3.416 & 974.536 \\
Llama-2-13B-chat & 68.190 & 2.039 & 10.141 & 3.332 & 1061.472 \\
Llama-2-70B-chat & 18.181 & 2.097 & 3.363 & 3.051 & 933.708 \\
Vicuna-13B & 20.587 & 2.463 & 3.858 & 3.082 & 891.109 \\
Orca-2-7B & 49.525 & 4.575 & 8.502 & 4.573 & 1042.146 \\
Orca-2-13B & 49.349 & 4.434 & 8.499 & 4.459 & 948.356 \\
\midrule
CodeLlama-7B & 67.783 & 3.993 & 15.805 & 15.995 & 4586.738 \\
CodeLlama-13B & 126.915 & 5.378 & 22.443 & 11.425 & 4037.748 \\
CodeLlama-34B & 70.241 & 4.720 & 13.151 & 8.122 & 2716.919 \\
CodeLlama-70B & 102.019 & 5.252 & 20.763 & 11.766 & 4692.574 \\
CodeLlama-7B-Instruct & 49.437 & 3.539 & 10.081 & 8.102 & 4469.528 \\
CodeLlama-13B-Instruct & 45.858 & 3.205 & 8.467 & 6.084 & 3802.495 \\
CodeLlama-34B-Instruct & 41.123 & 3.533 & 7.449 & 5.756 & 2915.341 \\
CodeLlama-70B-Instruct & 46.992 & 3.611 & 9.029 & 6.011 & 4658.893 \\
\midrule
Mistral-7B & 53.374 & 3.621 & 11.932 & 8.225 & 3890.356 \\
Mistral-7B-Instruct & 22.860 & 2.292 & 4.252 & 4.147 & 1199.339 \\
Openchat-3.5-7B & 38.451 & 2.515 & 6.748 & 5.220 & 4447.172 \\
Starling-LM-7B & 47.231 & 2.325 & 7.845 & 4.779 & 5483.139 \\
\midrule
GPT-3.5-0613 & 20.233 & 2.195 & 3.522 & 2.353 & 980.324 \\
GPT-3.5-1106 & 20.130 & 2.380 & 3.598 & 2.468 & 971.184 \\
GPT-4-0613 & 20.491 & 2.085 & 3.516 & 2.684 & 962.822 \\
GPT-4-1106 & 20.978 & 2.271 & 3.659 & 2.264 & 1423.798 \\

\bottomrule
\end{tabular}
\caption{Stack-Overflow Normal Setting (Mean Absolute Error)}
\label{tab:stack_overflow_normal_mae}
\end{table*}

\begin{table*}[hp]
\centering
\small
\begin{tabular}{lccrrr}
\toprule
Models & \begin{tabular}{c} FRE \end{tabular} & SMOG & FKGL & NERF & \begin{tabular}{c} Length \end{tabular} \\
\midrule
Human & 25.878 & 3.526 & 4.575 & 2.895 & 1243.833 \\
\midrule
Llama-2-7B & 137.635 & 4.339 & 29.784 & 17.822 & 4708.711 \\
Llama-2-13B & 139.464 & 5.575 & 35.837 & 63.870 & 4300.091 \\
Llama-2-70B & 123.341 & 6.373 & 25.888 & 20.171 & 3743.874 \\
\midrule
Llama-2-7B-chat & 17.229 & 1.953 & 3.131 & 3.396 & 1043.307 \\
Llama-2-13B-chat & 27.461 & 2.029 & 4.525 & 3.793 & 1031.519 \\
Llama-2-70B-chat & 16.896 & 2.020 & 3.125 & 3.024 & 952.276 \\
Vicuna-13B & 229.641 & 2.655 & 33.083 & 4.292 & 1006.485 \\
Orca-2-7B & 72.674 & 5.964 & 12.163 & 6.059 & 1260.131 \\
Orca-2-13B & 48.830 & 4.415 & 8.382 & 4.767 & 1097.196 \\
\midrule
CodeLlama-7B & 80.792 & 3.526 & 16.978 & 12.604 & 4844.746 \\
CodeLlama-13B & 91.405 & 4.852 & 17.564 & 8.963 & 3837.530 \\
CodeLlama-34B & 85.187 & 5.751 & 15.574 & 15.681 & 2673.364 \\
CodeLlama-70B & 114.259 & 5.776 & 23.174 & 12.558 & 4337.944 \\
CodeLlama-7B-Instruct & 41.407 & 2.943 & 8.409 & 8.134 & 5047.529 \\
CodeLlama-13B-Instruct & 54.775 & 3.954 & 10.138 & 7.931 & 4306.888 \\
CodeLlama-34B-Instruct & 29.485 & 2.501 & 4.989 & 4.976 & 1845.032 \\
CodeLlama-70B-Instruct & 40.356 & 3.577 & 7.526 & 6.220 & 5672.056 \\
\midrule
Mistral-7B & 42.017 & 3.016 & 9.444 & 7.830 & 3644.151 \\
Mistral-7B-Instruct & 21.777 & 2.086 & 3.847 & 4.080 & 1067.657 \\
Openchat-3.5-7B & 19.613 & 1.878 & 3.464 & 3.450 & 2332.179 \\
Starling-LM-7B & 28.696 & 2.450 & 4.505 & 4.134 & 6512.476 \\
\midrule
GPT-3.5-0613 & 21.340 & 2.627 & 3.721 & 2.558 & 981.147 \\
GPT-3.5-1106 & 25.569 & 3.976 & 4.691 & 2.625 & 934.750 \\
GPT-4-0613 & 23.365 & 3.316 & 4.194 & 2.435 & 979.349 \\
GPT-4-1106 & 25.152 & 4.068 & 4.766 & 2.576 & 1658.239 \\
\bottomrule
\end{tabular}
\caption{Stack-Overflow Complex Setting (Mean Absolute Error)}
\label{tab:stack_overflow_complex_mae}
\end{table*}

\begin{table*}[hp]
\centering
\small
\begin{tabular}{lrcrrr}
\toprule
Models & \begin{tabular}{c} FRE \end{tabular} & SMOG & FKGL & NERF & \begin{tabular}{c} Length \end{tabular} \\
\midrule
Human & 24.704 & 0.730 & 4.247 & 1.664 & 47.958 \\
\midrule
Llama-2-7B & 52.819 & 0.262 & 10.880 & 1.957 & 116.385 \\
Llama-2-13B & 46.716 & 0.201 & 9.104 & 8.925 & 169.677 \\
Llama-2-70B & 34.875 & 0.273 & 5.945 & 15.910 & 88.654 \\
\midrule
Llama-2-7B-chat & 30.699 & 0.968 & 5.071 & 1.913 & 67.696 \\
Llama-2-13B-chat & 87.097 & 1.992 & 13.059 & 2.426 & 345.292 \\
Llama-2-70B-chat & 29.152 & 1.489 & 4.882 & 8.678 & 104.558 \\
Vicuna-13B & 37.263 & 0.627 & 5.802 & 11.423 & 78.492 \\
Orca-2-7B & 32.298 & 0.517 & 5.305 & 1.902 & 50.550 \\
Orca-2-13B & 43.802 & 0.709 & 6.763 & 1.770 & 72.942 \\
\midrule
CodeLlama-7B & 69.643 & 0.249 & 13.698 & 4.857 & 95.962 \\
CodeLlama-13B & 77.903 & 0.276 & 12.222 & 5.135 & 104.465 \\
CodeLlama-34B & 62.395 & 0.271 & 9.858 & 25.323 & 114.354 \\
CodeLlama-70B & 67.085 & 0.249 & 12.484 & 19.832 & 197.304 \\
CodeLlama-7B-Instruct & 36.872 & 0.378 & 5.936 & 1.624 & 66.681 \\
CodeLlama-13B-Instruct & 54.860 & 0.259 & 11.146 & 13.875 & 126.096 \\
CodeLlama-34B-Instruct & 39.097 & 0.273 & 6.500 & 45.192 & 222.504 \\
CodeLlama-70B-Instruct & 47.679 & 0.299 & 10.115 & 16.811 & 173.119 \\
\midrule
Mistral-7B & 47.280 & 0.205 & 9.545 & 5.077 & 119.508 \\
Mistral-7B-Instruct & 34.043 & 0.277 & 6.084 & 18.761 & 207.588 \\
Openchat-3.5-7B & 30.908 & 1.580 & 5.260 & 5.334 & 525.646 \\
Starling-LM-7B & 34.648 & 7.653 & 6.012 & 3.222 & 5062.912 \\
\midrule
GPT-3.5-0613 & 23.879 & 6.412 & 4.047 & 2.916 & 160.192 \\
GPT-3.5-1106 & 24.334 & 4.758 & 4.170 & 2.740 & 108.650 \\
GPT-4-0613 & 24.354 & 4.991 & 4.170 & 2.807 & 111.985 \\
GPT-4-1106 & 24.288 & 7.606 & 4.270 & 3.371 & 212.377 \\
\bottomrule
\end{tabular}
\caption{TSCC Setting (Mean Absolute Error)}
\label{tab:tscc_mae}
\end{table*}

\subsection{Skip rows}
Table \ref{tab:skip_rows} presents the skipped rows. As indicated in Table \ref{tab:skip_rows}, instruction-tuned models adhere to the formats, exhibiting only a few skipped rows, with the exception of CodeLLaMA.

\subsection{Case Study}
Table \ref{tab:examples_dialogues} presents the extraction of a single-turn teacher response for evaluation. As illustrated in Table \ref{tab:examples_dialogues}, we compare the utterance "with hail and everything" to the response "Ooh, I hope you’re not too badly affected by them!" focusing on text difficulty, synonymity, and redundancy.

\begin{table*}[hp]
\centering
\small
\begin{tabular}{lcccc}
\toprule
Models & \multicolumn{3}{c}{Stack-Overflow} & TSCC \\
\midrule
Settings & normal & simple & complex & -- \\
\midrule[0.02mm]
Human & 0 & 0 & 0 & 0 \\
\midrule[0.25mm]
Llama-2-7B & 16 & 15 & 14 & 0 \\
Llama-2-13B & 7 & 6 & 6 & 4 \\
Llama-2-70B & 16 & 16 & 16 & 3 \\
\midrule
Llama-2-7B-chat & 0 & 0 & 0 & 0 \\
Llama-2-13B-chat & 0 & 0 & 0 & 0 \\
Llama-2-70B-chat & 0 & 0 & 0 & 2 \\
Vicuna-13B & 0 & 0 & 0 & 5 \\
Orca-2-7B & 0 & 3 & 0 & 1 \\
Orca-2-13B & 0 & 0 & 0 & 1 \\
\midrule
CodeLlama-7B & 16 & 16 & 16 & 4 \\
CodeLlama-13B & 16 & 16 & 16 & 5 \\
CodeLlama-34B & 16 & 16 & 16 & 5 \\
CodeLlama-70B & 16 & 16 & 16 & 5 \\
CodeLlama-7B-Instruct & 15 & 13 & 14 & 4 \\
CodeLlama-13B-Instruct & 15 & 16 & 16 & 4 \\
CodeLlama-34B-Instruct & 16 & 16 & 16 & 4 \\
CodeLlama-70B-Instruct & 13 & 15 & 16 & 2 \\
\midrule
Mistral-7B & 15 & 15 & 15 & 1 \\
Mistral-7B-Instruct & 1 & 0 & 0 & 4 \\
Openchat-3.5-7B & 0 & 0 & 0 & 0 \\
Starling-LM-7B & 0 & 0 & 0 & 0 \\
\midrule
GPT-3.5-0613 & 0 & 0 & 0 & 0 \\
GPT-3.5-1106 & 0 & 0 & 0 & 0 \\
GPT-4-0613 & 0 & 0 & 0 & 0 \\
GPT-4-1106 & 0 & 0 & 0 & 0 \\
\bottomrule
\end{tabular}
\caption{Skip rows}
\label{tab:skip_rows}
\end{table*}

\begin{sidewaystable*}[hp]
\centering
\small
\begin{tabular}{lllllll}
\toprule
Models & Base Models & Parameter Size & Datasets & Tuning Methods & versions & Author \\
\midrule
Llama-2 & (Base) & 7B, 13B, 70B & Publicly available sources & (Base) & Llama-2-\{7,13,70\}b-hf & \cite{touvron2023llama2} \\
Llama-2-chat & Llama-2 & 7B, 13B, 70B & Publicly available sources & SFT + RLFT & Llama-2-\{7,13,70\}b-chat-hf & \cite{touvron2023llama2} \\
Vicuna & Llama-2 & 13B & ShareGPT & SFT & vicuna-13b-v1.5 & \cite{zheng2023judging} \\
Orca & Llama-2 & 7B, 13B & Publicly available sources & SFT & microsoft/Orca-2-\{7,13\}b & \cite{mitra2023orca} \\
CodeLlama & Llama-2 & 7B, 13B, 34B, 70B & Publicly available sources & SFT + RLFT & CodeLlama-\{7,13,34,70\}b-hf & \cite{roziere2023code} \\
CodeLlama-Instruct & Llama-2 & 7B, 13B, 34B, 70B & Publicly available sources & SFT + RLFT & CodeLlama-\{7,13,34,70\}b-Instruct-hf & \cite{roziere2023code} \\
Mistral & (Base) & 7B & -- & (Base) & Mistral-7B-v0.1 & \cite{jiang2023mistral} \\
Mistral-Instruct & Mistral & 7B & -- & SFT & Mistral-7B-Instruct-v0.1 & \cite{jiang2023mistral} \\
Openchat & Mistral & 7B & ShareGPT & C-RLFT & openchat\_3.5 & \cite{wang2023openchat} \\
Starling & Openchat & 7B & Nector & C-RLFT+ APA & Starling-LM-7B-alpha & \cite{starling2023} \\
GPT3.5-Turbo & -- & -- & -- & SFT + RLFT & gpt-3.5-turbo-\{0613, 1106\} & \cite{ouyang2022training} \\
GPT4 & -- & -- & -- & SFT + RLFT & gpt-4-0613 & \cite{openai2023gpt4} \\
GPT4-Turbo & -- & -- & -- & -- & gpt-4-1106-preview & \cite{openai2023gpt4} \\
\bottomrule
\end{tabular}
\caption{Models Description}
\label{tab:models_description}
\end{sidewaystable*}

\section{Models Description}
\label{sec:models_description}

\paragraph{ChatGPT} \cite{ouyang2022training, openai2023gpt4} is an LLM that employs Reinforcement Learning from Human Feedback (RLHF) to align with human preferences, and it stands out for its exceptionally high performance among current language models.

\paragraph{LLaMA-2} \cite{touvron2023llama2} is an LLM pre-trained and fine-tuned across a range of 700 million to 7 billion parameters. This model not only outperforms LLaMA and its variants \cite{touvron2023llama} in numerous benchmarks but has also undergone manual reviews for its usefulness and safety, indicating its potential to substitute closed-source models. Besides, it includes variations with different parameter sizes and versions fine-tuned for dialogue data and source code, such as LLaMA-2-chat and Code-LLaMA \cite{roziere2023code}.

\paragraph{Vicuna} \cite{zheng2023judging} is an LLM trained to align with human preferences using data from ShareGPT \footnote{\url{https://sharegpt.com/}} interactions, and based on LLaMA \cite{touvron2023llama}. 
We selected the 1.5 version of this model based on LLaMA-2 to analyze the impact of on text difficulty adaptation.

\paragraph{Orca} \cite{mitra2023orca} is a model fine-tuned with prompts from various strategies, enabling it to adjust difficulty and offer flexible outputs in response to input sentences.

\paragraph{Mistral} \cite{jiang2023mistral} is a pre-trained model with 7 billion parameters. Compared to the larger parameter-sized 13B model of LLaMA-2, Mistral has recorded high performance in benchmarks.

\paragraph{OpenChat} \cite{wang2023openchat} builds on Mistral \cite{jiang2023mistral} and ShareGPT for training, enhancing learning by leveraging data quality variance between GPT-3.5 and GPT-4 as a reward mechanism.

\paragraph{Starling} \cite{starling2023} is trained with a reward model derived from feedback on GPT-4 \cite{openai2023gpt4} and builds upon OpenChat \cite{wang2023openchat}, which itself was fine-tuned from Mistral. We aim to explore whether models based on Mistral can develop the ability to modulate difficulty levels through fine-tuning.

Table \ref{tab:models_description} shows various training methods for model tuning, including Supervised Fine-Tuning (SFT \cite{xu2023wizardlm, ding-etal-2023-enhancing}), Reinforcement Learning Fine-Tuning (RLFT)~\cite{schulman2017proximal, ouyang2022training}, Conditioned RLFT (C-RLFT)~ \cite{wang2023openchat}, Advantage-Induced Policy Alignment (APA)~\cite{zhu2023fine}, and Direct Preference Optimization (DPO)~\cite{rafailov2023direct}.


\section{Dataset Construction}
\label{sec:appendix:dataset_construction}
We construct a dataset for effectively comparing text difficulty, which consists of two parts, questions and answers. Both components feature sentences of sufficient length to ensure accurate difficulty estimation. Since short target sentences can lead to potentially inaccurate difficulty assessments, existing QA datasets such as SQuAD~\cite{rajpurkar-etal-2016-squad}, which typically contain brief answers (for example, a single word or sentence), do not meet our criteria.

To address this challenge, we created a dataset from Stack-Overflow, selecting data as of July 1, 2023. Considering the extended text lengths within the collected dataset, we extracted 1,000 posts in a novel order to optimize the scope of feasible experiments under constrained resources. The extracted posts contain significantly more tokens than typically observed in QA datasets, as detailed in Appendix \ref{sec:appendix:long_input}.

We then extracted the "QuestionTitle," "QuestionBody," and "AnswerBody" fields from each post. We combined "QuestionTitle" and "QuestionBody" to form the Questions component and designated "AnswerBody" as the Answers.
We will release our code and dataset at \url{https://github.com/satoshi-2000/llms-suitable}.

\section{Packages}
We used several packages for scoring such as evaluate ($ver.\ 0.4.0$)\footnote{https://huggingface.co/docs/evaluate/index}, textstat ($ver.\ 0.7.3$) \footnote{https://github.com/textstat/textstat}, spacy ($ver.\ 3.5.2$) \footnote{https://spacy.io/}, and lftk ($ver.\ 1.0.9$) \footnote{https://github.com/brucewlee/lftk}.

\section{
Ensuring License Compliance in Artifact Usage}
We reviewed the license terms before comparing models to ensure adherence to the intended use.
Additionally, we utilized AI assistants, including ChatGPT and Copilot, for coding and writing the thesis.

\end{document}